%% file: goparaju_shapemi_2018.tex
\documentclass[runningheads]{llncs}
 \usepackage{float}
\usepackage{placeins}
\usepackage{graphicx}
\usepackage{tikz}
\def\checkmark{\tikz\fill[scale=0.4](0,.35) -- (.25,0) -- (1,.7) -- (.25,.15) -- cycle;} 

\def\xmark{{\sffamily X}} 

\usepackage{booktabs}

\usepackage[font=small,skip=0pt]{caption}

\usepackage{amsfonts}
\newcommand{\mcS}{\mathcal{S}}

\newcommand{\mbbR}{\mathbb{R}}

\newcommand{\mbfz}{\mathbf{z}}
\newcommand{\mbfZ}{\mathbf{Z}}
\newcommand{\mbfx}{\mathbf{x}}
\newcommand{\mbfX}{\mathbf{X}}
\newcommand{\mbfT}{\mathbf{T}}

\begin{document}
%
%\title{On the Evaluation, Validation, and Selection of Off-the-shelf Statistical Shape Modeling Tools: \\A Clinical Application}
\title{On the Evaluation and Validation of Off-the-shelf Statistical Shape Modeling Tools: \\
	A Clinical Application}

\titlerunning{Evaluation and Validation of SSM Tools}

\author{Anupama Goparaju\inst{1} \and 
			Ibolya Csecs\inst{2} \and
			Alan Morris\inst{2} \and 
			Evgueni Kholmovski\inst{2,3} \and 
			Nassir Marrouche\inst{2} \and 
			Ross Whitaker\inst{1} \and 
			Shireen Elhabian\inst{1}}
	
%\author{First Author\inst{1}\orcidID{0000-1111-2222-3333} \and
%Second Author\inst{2,3}\orcidID{1111-2222-3333-4444} \and
%Third Author\inst{3}\orcidID{2222--3333-4444-5555}}
%
\authorrunning{Goparaju et al.}

\institute{
	Scientific Computing and Imaging Institute, University of Utah, SLC, UT, USA 
	\email{anupama.goparaju@utah.edu},
	\email{\{whitaker,shireen\}@sci.utah.edu}\\
	\and
	Comprehensive Arrhythmia Research and Management Center, Division of Cardiovascular Medicine, School of Medicine, University of Utah,  SLC, UT, USA 
	\email{\{alan.morris,nassir.marrouche\}@carma.utah.edu}
	\and
 	Department of Radiology and Imaging Sciences, School of Medicine, University of Utah, SLC, UT, USA 
	\email{evgueni.kholmovski@hsc.utah.edu}
}

%\institute{Princeton University, Princeton NJ 08544, USA \and
%Springer Heidelberg, Tiergartenstr. 17, 69121 Heidelberg, Germany

\maketitle              % typeset the header of the contribution

\input{abstract}

\input{introduction}

\input{methods}
\input{results}

\input{conclusion}

\vspace{0.05in}
\noindent\textbf{Acknowledgment:} This work was supported by NIH %the National Institutes of Health 
[grant numbers R01-HL135568-01 and P41-GM103545-19] and Coherex Medical. 

\vspace{-0.1in}
{\small 
\bibliographystyle{splncs04}
\bibliography{references}
}
\end{document}

%% file: abstract.tex
\begin{abstract}
Statistical shape modeling (SSM) has proven useful in many areas of biology and medicine as a new generation of morphometric approaches for the quantitative analysis of anatomical shapes. Recently, the increased availability of high-resolution in vivo images of anatomy has led to the development and distribution of open-source computational tools to model anatomical shapes and their variability within populations with unprecedented detail and statistical power. 
Nonetheless, there is little work on the evaluation and validation of such tools as related to clinical applications that rely on morphometric quantifications for treatment planning. 
To address this lack of validation, we systematically assess the outcome of widely used off-the-shelf SSM tools, namely ShapeWorks, SPHARM-PDM, and Deformetrica,  in the context of designing closure devices for left atrium appendage (LAA) in atrial fibrillation (AF) patients to prevent stroke, where an incomplete LAA closure may be worse than no closure. 
This study is motivated by the potential role of SSM in the geometric design of closure devices, which could be informed by population-level statistics,  and patient-specific device selection, which is driven by anatomical measurements that could be automated by relating patient-level anatomy to population-level morphometrics. 
Hence, understanding the consequences of different SSM tools for the final analysis is critical for the careful choice of the tool to be deployed in real clinical scenarios. 
Results demonstrate that estimated measurements from ShapeWorks model are more consistent compared to models from Deformetrica and SPHARM-PDM. Furthermore, ShapeWorks and Deformetrica shape models capture clinically relevant population-level variability compared to SPHARM-PDM models. 
	
\keywords{Statistical shape models  \and Surface parameterization \and Correspondence optimization \and Evaluation.}
\end{abstract}

%% file: introduction.tex
\section{Introduction}
\vspace{-0.1in}

Morphometric techniques for the quantitative analysis of anatomical shapes have been important for the study of biology and medicine for more than 100 years.
Statistical shape modeling (SSM) is the computational extension of classical morphometric techniques to more detailed representations of complex anatomy and their variability within populations with high levels of geometric detail and statistical power. 
SSM is beginning to impact a wide spectrum of clinical applications, e.g., implants design \cite{zadpoor2015patient}, anatomy reconstruction from less-expensive 2D images \cite{carlier2015bringing}, surgical planning \cite{markelj2012review}, and reconstructive surgery \cite{zachow2015computational}. 

\vspace{0.04in}
\noindent\textbf{Learning population-level metric:} Developing computational tools for shape modeling is contingent upon defining a \textit{metric} in the space of shapes to enable comparing shapes and performing shape statistics (e.g., averaging). That is, two shapes that differ in a manner that is typical of the variability in the population should be considered \textit{similar} relative to two shapes that differ in atypical ways. For instance, size is such a typical mode of anatomical variation that most shape-based analyses factor it out, thereby treating two anatomical objects that differ only in size as the \textit{same}. Populations of anatomic objects typically show other common variations. 
There is a growing consensus in the field that such a metric should be adapted to the specific population under investigation, which entails finding correspondences across an ensemble of shapes. The scientific premise of existing correspondence techniques falls in two broad categories: a \textit{groupwise} approach to estimating 
correspondences (e.g., ShapeWorks \cite{cates2017shapeworks}, Minimum Description Length - MDL \cite{davies2002learning}, Deformetrica \cite{durrleman2014morphometry}) that considers the variability in the entire cohort 
and a \textit{pairwise} approach (e.g., SPHARM-PDM \cite{styner2006framework}) 
that considers mapping to a predefined surface parameterization. Pairwise methods lead to biased and suboptimal 
models \cite{oguz2015entropy,dalal2010multiple,davies2008statistical}. On the other hand, groupwise methods learn a population-specific metric in a way that does not penalize natural variability and therefore can capture the underlying parameters in an anatomical shape space. Other publicly available tools, e.g., FreeSurfer \cite{fischl1999high}, BrainVoyager \cite{goebel2006analysis}, FSL \cite{jenkinson2012fsl}, and SPM \cite{ashburner2012spm}, provide shape modeling capabilities, but they tend to be tailored to specific anatomies or limited topologies. SPHARM-PDM  
\cite{styner2006framework}, for example, is a parameterization-based correspondence scheme that relies on a smooth one-to-one mapping from each surface instance to the unit sphere. Here, we consider a representative set of open-source SSM tools (see Table \ref{tab:ssm_tools}) that can be used for general anatomy; 
ShapeWorks, Deformetrica, and SPHARM-PDM. 

% --------------- table of comparison -----------------
\begin{table}
	\caption{Open-source SSM tools considered for evaluation and validation \vspace{-0.015in}}
	\label{tab:ssm_tools}
	\centering
	{\footnotesize
	\resizebox{\linewidth}{!}{%
		\begin{tabular}{p{0.25\linewidth}p{0.15\linewidth}p{0.3\linewidth}p{0.33\linewidth}p{0.27\linewidth}} \toprule 
			\textbf{SSM tools} &\textbf{Groupwise}   & \textbf{Topology-independent} & \textbf{%Surface
				Parameterization-free} & \textbf{General anatomy} \\ \toprule
			ShapeWorks \cite{cates2017shapeworks} & \checkmark & \checkmark & \checkmark& \checkmark \\ \midrule 
			SPHARM-PDM \cite{styner2006framework} & \xmark & \xmark & \xmark  (sphere)& \checkmark (spherical topology)  \\ \midrule 
			Deformetrica \cite{durrleman2014morphometry} & \checkmark & \checkmark & \checkmark & \checkmark   \\ \midrule 
	\end{tabular}}\vspace*{-0.5\baselineskip}
	}
\end{table}

\begin{figure}
	\centering%\vspace{-0.2in}
	\includegraphics[width=0.8\linewidth]{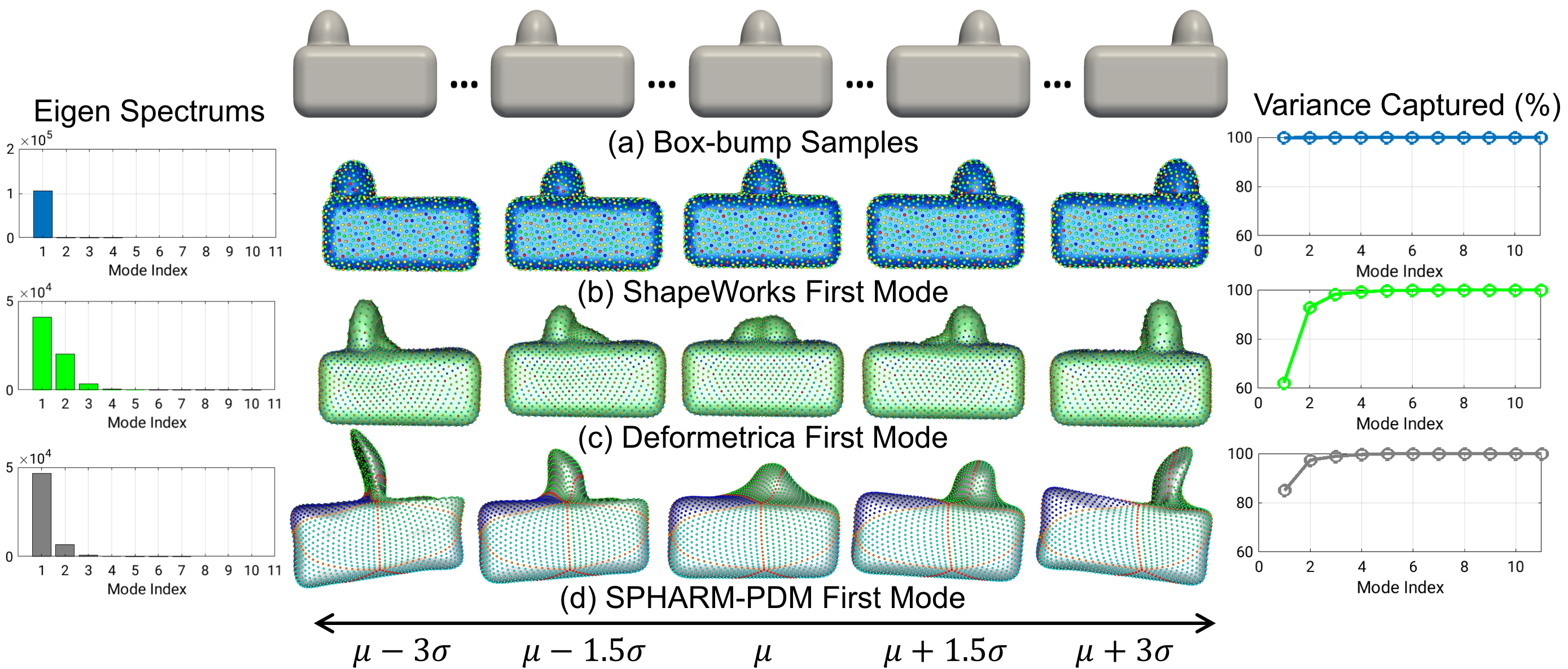}
	\caption{{\small \textbf{Proof-of-concept:} (a) Box-bump samples. The mean $\pm$ 3 stds of the 1st mode of (b) ShapeWorks \cite{cates2017shapeworks}, (c) Deformetrica \cite{durrleman2014morphometry}, and (d) SPHARM-PDM \cite{styner2006framework}.}}\vspace{0.07in}
	\label{fig:box_bump}
\end{figure}

% box-bump experiment to motivate the need for evaluation and validation
\vspace{0.04in}
\noindent\textbf{Proof-of-concept:} %As a demonstration, 
Consider the simple example of an ensemble of 3D boxes with a bump at a varying location. Ideally, one would want the correspondences to reflect the fact that the {\em bump} is a single feature whose position on the main box shape varies across the population. When comparing the shape of different boxes, one would want to downplay the impact of the bump location on the comparing metric to respect the natural population variability that is not captured by simple affine transformations. Figure \ref{fig:box_bump} illustrates how the groupwise aspect of the corresponence optimization of ShapeWorks is able to discover the underlying mode of variation in the box-bump ensemble in comparison to a pairwise diffeomorphism-based %(coordinate transformation) 
shape modeling approach (Figure \ref{fig:box_bump}(c)) in which shapes are embedded in the image intensity values at voxels, and nonlinear registration is used to map all sample images to a reference image, which is estimated based on the Frechet mean of all samples \cite{durrleman2014morphometry}. As illustrated in Figure \ref{fig:box_bump}(a), shapes from ShapeWorks remain more faithful to those described by the original training set, even out to three standard deviations, at which the diffeomorphic description breaks down. In particular, diffeomorphic warps recovered an incorrect shape model in which the mean shape showed a box with two bumps rather than a single bump. Furthermore, SPHARM-PDM does not guarantee an efficient solution in the parameter space of the resulting shape model (Figure \ref{fig:box_bump}(d)), with an inherent limitation of only modeling anatomies with spherical topology.

\vspace{0.04in}
\noindent\textbf{Lack of validation in a clinical scenario:} Computer-assisted diagnosis and surgical planning can help clinicans making 
objective decisions \cite{shinya2011,naiara2017ssm}. In particular, shape modeling has played an important role in clinical applications that benefit from both qualitative and quantitative insights of population-level variability, e.g., diagnosis of liver cirrhosis \cite{shinya2011} and finding associations between surgical parameters and head shapes following cranioplasty \cite{naiara2017ssm}. 
Recent advances in in vivo imaging of anatomy and the wide spectrum of shape modeling applications have led to the development and distribution of open-source SSM tools, further enabling their use in an \textit{off-the-shelf} manner. 
Evaluation of SSM tools has been performed using non-clinical applications such as image segmentation \cite {gollmer2014} and shape/deformation synthesis approaches \cite{gao2014}. 
However, to the best of our knowledge, little work has been done on the evaluation and validation of such tools as related to clinical applications that rely on morphometric quantifications. Specifically, we believe that understanding the consequences of different SSM tools on the final analysis is critical for the careful choice of the tool to be deployed in real clinical scenarios.
To address this lack of validation in a clinical scenario, we systematically assess the outcome of widely used off-the-shelf SSM tools, namely ShapeWorks \cite{cates2017shapeworks}, SPHARM-PDM \cite{styner2006framework}, and Deformetrica \cite{durrleman2014morphometry},  in the context of designing closure devices for left atrium appendage (LAA) in atrial fibrillation (AF) patients to prevent stroke. 

\vspace{0.04in}
\noindent\textbf{Closure device design -- a potential clinical application:} 
LAA closure is performed in AF patients to reduce the risk of stroke \cite{wang2010}. LAA morphology is complex and mainly divided into four types: cauliflower, chicken wing, wind sock, and cactus \cite{wang2010}.  The geometric design and patient-specific selection of closure devices are typically performed by clinical experts with subjective decisions based on the morphology \cite{wang2010}.  Nonetheless, obtaining these measurements manually for large cohorts of patients is a subjective, tedious, and error-prone process. SSM could thus provide an automated approach for developing less subjective categorizations of LAA morphology and measurements that can be used to make more objective clinical decisions regarding suitability for LAA closure. 
Hence, this study is motivated by the potential role of SSM in the geometric design of closure devices, which could be informed by population-level statistics,  and patient-specific device selection, which is driven by anatomical measurements that could be automated by relating patient-level anatomy to population-level morphometrics.
To validate different SSM tools, we present a semiautomated approach that makes use of shape models to estimate the LAA measurements to aid the patient-specific closure device selection process; a clinical application that needs consistent and accurate measurement estimation to avoid adverse outcomes that could result from incomplete appendage closure \cite{regazzoli2015left}.

%% file: methods.tex
\vspace{-0.1in}
\section{Methods}
\vspace{-0.1in}

The crux of statistical shape modeling is defining a shape representation that allows performing shape statistics. To this end, \textit{landmarks} are the most popular choice as a light-weight shape representation that is easy to understand and that promotes visual communication of the results \cite{sarkalkan2014statistical,zachow2015computational}. To perform shape arithmetics (e.g., averaging), landmarks should be defined consistently within a given population to refer to the same anatomical position on every shape instance, a concept known as \textit{correspondence}.  
Given an ensemble of shapes for a particular anatomy, these correspondences (or landmarks/points ) are typically generated using some optimization process by defining an objective function to be minimized in a \textit{pairwise} (w.r.t. a shape atlas/template or predefined surface parameterization, e.g., sphere) or \textit{groupwise} (w.r.t. all shape samples where each shape provides new information about shape variability) manner. 
Different SSM tools implement different objective functions \cite{gollmer2014}, raising the need to evaluate and validate their resulting shape models in clinical applications that rely on shape-based measurements. %To this regard, 
As such, we present a semiautomated approach to validate the results of widely used SSM tools in one such clinical application. 

\vspace{-0.15in}
\subsection{Statistical shape models}
\vspace{-0.05in}

Statistical shape models consist of (1) a detailed 3D geometrical representation of the \textit{average anatomy} of a given population and (2) a representation of the population-level \textit{geometric variability} of such an anatomy, often in the form of a collection of principal modes of variation. Principal modes of variation define reduced degrees of freedom for representing a high-dimensional and thus complex variation in anatomical shapes. In particular, consider a cohort of shapes $\mcS = \{\mbfz_1, \mbfz_2, ..., \mbfz_N\}$ of $N$ surfaces, each with its own set of $M$ corresponding point $\mbfz_n = [\mbfz_n^1, \mbfz_n^2, ..., \mbfz_n^M] \in \mbbR^{dM}$ where the ordering of each point $\mbfz_n^m \in \mbbR^d$ implies a correspondence among shapes. For statistical modeling, shapes in $\mcS$ should share the same world coordinate system. Hence, a rigid transformation matrix $\mbfT_n$  can be estimated to transform the points in the $n-$th shape local coordinate $\mbfx_n^m$ to the world common coordinate $\mbfz_n^m$ such that $\mbfz_n^m = \mbfT_n\mbfx_n^m$. 
Using principal component analysis (PCA), this high-dimensional point distribution model (PDM) can be reduced to a compact set of $K$ modes of variations.

%m
\vspace{-0.15in}
\subsection{SSM tools}
\vspace{-0.05in}

Here is a brief summary of the SSM tools considered in this study.

\vspace{0.04in}
\noindent\textbf{ShapeWorks \cite{cates2017shapeworks}}
is a groupwise correspondence approach that implements the particle-based modeling method (PBM) \cite{cates2007shape}, which constructs compact statistical landmark-based models of shape ensembles that do not rely on any specific surface parameterization. 
It uses a set of interacting particle systems, one for each shape, using mutually repelling forces to optimally cover, and therefore describe, the surface geometry, thus avoiding many of the problems inherent in parametric representations, such as the limitation to specific topologies, processing steps necessary to construct parameterizations, and bias toward model initialization.
PBM considers two types of random variables: a shape space variable $\mbfZ \in \mbbR^{dM}$ and a particle position variable $\mbfX_n \in \mbbR^d$ that encodes particles distribution on the $n-$th shape (configuration space). Correspondences are established by minimizing a combined shape correspondence and surface sampling cost function $Q = H(\mbfZ) - \sum_{n=1}^N H(\mbfX_n)$,  where $H$ is an entropy estimation assuming Gaussian shape distribution in the shape space and Euclidean particle-to-particle repulsion in the configuration space. This formulation favors a compact ensemble representation in shape space (first term) against a uniform distribution of particles on each surface for accurate shape representation (second term).

\vspace{0.04in}
\noindent\textbf{SPHARM-PDM \cite{styner2006framework}} is a pairwise and parameterization-based correspondence method that maps each training sample to a unit sphere with an area preserving and distortion minimizing objective using spherical harmonics as basis functions. Hence, SPHARM is restricted to anatomies with spherical topology. 
Spherical parameterization is obtained by aligning the axes of  first order ellipsoid fit of an input shape to the basis functions axes, 
The basis function of degree $l$ and order $m$ is given as, ${Y_l}^m(\theta \phi)=\sqrt{\frac{2l+1}{4\pi} \frac{(l-m)!}{(l+m)!}}{P_l}^m(\cos\theta)e^{im\phi}$, where $\theta \in [0;\pi]$ and $\phi \in [0;2\pi]$. Every point on the surface is given by a parameter vector ($\theta_i$, $\phi_i$), which represents a location on the sphere. Each mesh's spherical parameterization is used in generation of the SPHARM description. Icosahedron subdivision of a sphere 
is performed to obtain homogeneous sampling of the parameter space and thereby obtain a point distribution model (PDM). 

\vspace{0.04in}
\noindent\textbf{Deformetrica \cite{durrleman2014morphometry}} is a deformation-based correspondence method that is based on the large deformation diffeomorphic metric mapping (LDDMM) framework. A deformation field $X(x)$ is generated using $n$ control points ($q_i)_{i=1,2..,n} $ and momenta vectors ($\mu_i)_{i=1,2..,n}$, where $X(x)=\sum_{i=1}^{p} K(x,q_i)\mu _i$, $x$ is the position at which the vector field is evaluated, and $K(x,y)$ is a Gaussian kernel with width $\sigma$. The deformation is obtained through the convolution between the control points and their momenta and the vertices of the input meshes. Varifold distance is used in estimating the distance between meshes. The transformation obtained by the deformation is denoted as $\phi_{q,\mu}$, where $q$ and $\mu$ are the initial control points and momenta. 
The number of control points and the topology of the atlas are user-defined. The algorithm is initialized with the control points on the atlas and momenta vectors set to zero indicating no deformation. A path of deformations is estimated by mapping the atlas to the input shape. A final atlas is generated with the optimized control points and momenta vectors by considering the variability of all the input shapes. The deformations inform how different a shape is from the atlas and enable statistical shape analysis.

\vspace{-0.15in}
\subsection{Evaluation methodology}
\vspace{-0.05in}

When the ground truth data of shape descriptors is not available, evaluation of shape models can be performed using \textit{quantitative} metrics reported as a function of the number of principal modes of variation, namely \cite{davies20023d}: (1) \textit{compactness}, which encodes the percentage of variance captured by a specific number of modes (higher is better), and thus a compact shape model would express the shape variability with fewer parameters; (2) \textit{generalization}, which assesses whether a learned model can represent unseen shape instances (using reconstruction error in a leave-one-out cross validation scheme) and quantifies the ability of the learned density function to spread out between and around the training shapes (lower is better); and (3) \textit{specificity}, which reveals the ability of the model to generate plausible/realistic shapes, quantified as the Euclidean distance between a sampled shape and its closest training sample based on $\ell_2$ norm (lower is better).

A \textit{qualitative} analysis can be performed by analyzing the mean (average) anatomy and learned modes of variation. 
Furthermore, when the given population exhibits natural clustering (e.g., LAA), SSM tools can be evaluated by performing clustering analysis on the resulting correspondences to analyze which tool can help discover the intrinsic variability in the data.

\vspace{-0.15in}
\subsection{Validation methodology}
\vspace{-0.05in}

Shape models for different SSM tools can be validated using manually defined landmarks. However, defining manual correspondences is time-consuming, especially when dealing with large cohorts, and requires significant human resources with sufficient domain-specific expertise. 
To address the issue of validation with the ground truth correspondences, we have designed a semiautomated approach to test the robustness of shape models in the context of the LAA closure process. 
Closure devices are available in different sizes and device design and selection is made by a clinician using manual analysis of a patient's LAA and the ostium (opening of LAA) \cite{wang2010}. 
The manual analysis can be automated by estimating the anatomical measurements of patient-specific LAA by relating it to the population-level statistics informed by shape models. The anatomical measurement estimation process is performed by (1) marking the correspondences of LAA ostium on the mean shape obtained from the shape model of an SSM tool, (2) warping the marked ostium to the individual samples (using mean-sample correspondences to construct a thin-plate spline (TPS) warp \cite{bookstein1989principal}) to obtain the geometry of a single sample LAA ostium, and (3) computing anatomical measurements using sample-specific ostium points.  
Measurements obtained by this SSM-based semiautomated process are then compared against ground truth measurements (defined by clinical experts) to estimate the accuracy of correspondences established by an SSM tool.

%% file: results.tex
\vspace{-0.2in}
\section{Results}
\vspace{-0.05in}

\vspace{-0.1in}
\subsection{Experimental setup}
\vspace{-0.05in} 

SSM tools were evaluated and validated using a dataset of 130 LAA (isotropic resolution of 0.625 mm) that was obtained retrospectively from an AF patient image database at the University of Utah's Comprehensive Arrhythmia Research and Management (CARMA) Center. The MRI volumes were served with a single-handed segmentation by an expert. A preprocessing pipeline of rigid registration to a representative sample to factor out translational and rotational variations, cropping (using a largest bounding box of the dataset) to remove the unnecessary background that could slow down the processing, and topology-preserving smoothing were applied to the LAA shapes to generate signed distance transforms. 
The preprocessed shapes were then fed to ShapeWorks, SPHARM-PDM, Deformetrica with a mean LAA shape (distance tranform -- $\mu$DT) as an atlas, and Deformetrica with a sphere shape as an atlas (similar to SPHARM-PDM). 
Resulting 3D point correspondences from different SSM tools were then used for evaluation and validation of the tools. 

\begin{figure}[!htp]
	\centering%\vspace{-0.2in}
	\includegraphics[width=0.85\linewidth]{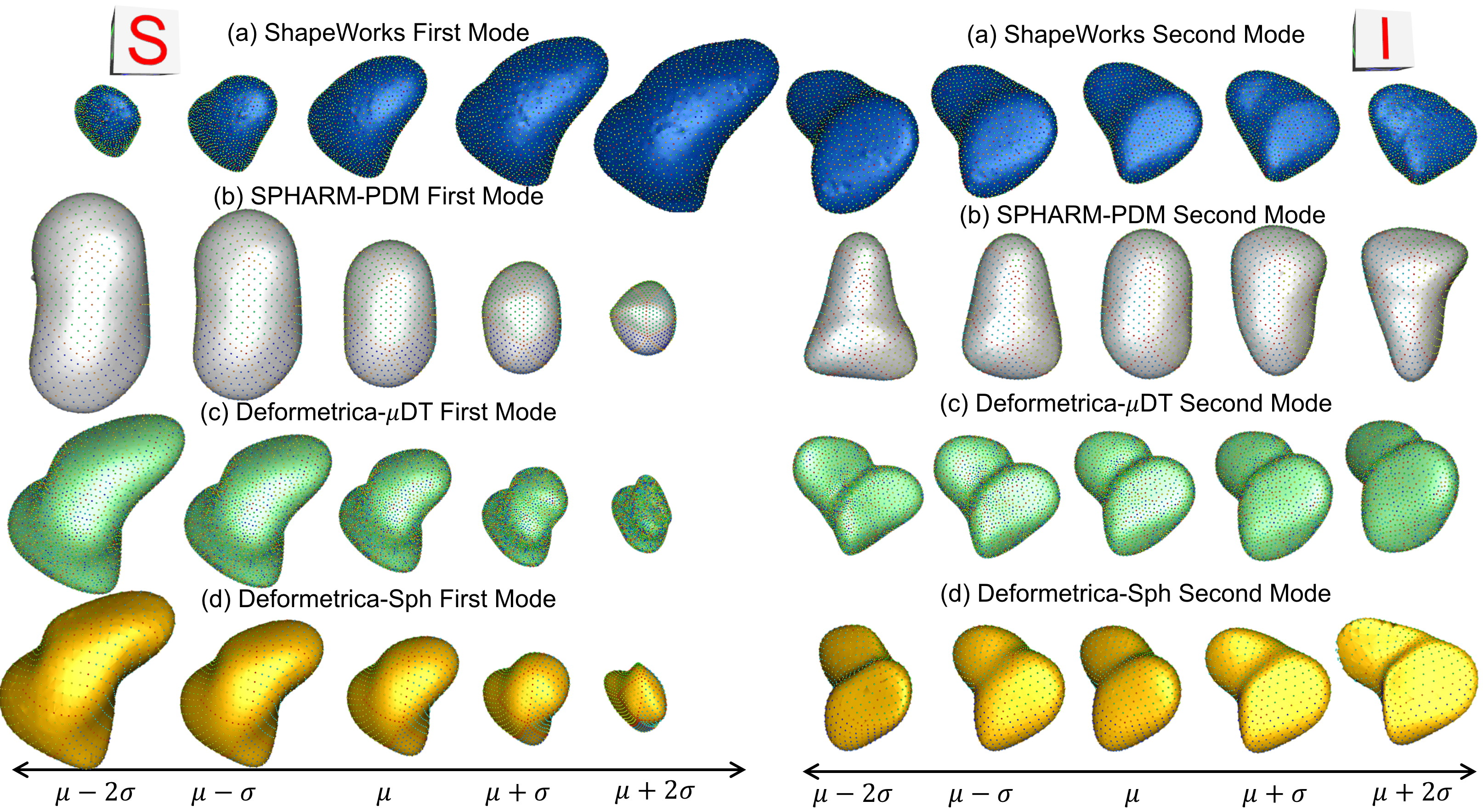}
		%camera ready addition below : (Superior and Inferior Views)
	\caption{The mean $\pm$ 2 stds of the 1st and 2nd mode of LAA shape models (S: superior and I: inferior views) from (a) ShapeWorks \cite{cates2017shapeworks}, (b) SPHARM-PDM \cite{styner2006framework}, (c) Deformetrica-$\mu$DT (mean LAA distance transform as the atlas) \cite{durrleman2014morphometry}, and (d) Deformetrica-Sph (sphere shape as the atlas) \cite{durrleman2014morphometry}.}
	\label{fig:modes}
\end{figure}

\vspace{-0.2in}
\subsection{Shape models evaluation}
\vspace{-0.05in} 

\noindent\textbf{Modes of variation: }Figure \ref{fig:modes} illustrates the first two dominant modes from different tools. In contrast to SPHARM-PDM, shape models from ShapeWorks and Deformetrica capture clinically relevant variations: the elongation of the appendage and the size of LAA ostia. Of particular interest, the second dominant mode of variation from SPHARM-PDM reflects the ambiguity in axes mapping of first order ellipsoid fit of the input shape to the axes in the parameter space.

\begin{figure}[!htp]
	\centering%\vspace{-0.2in}
	\includegraphics[width=0.9\linewidth]{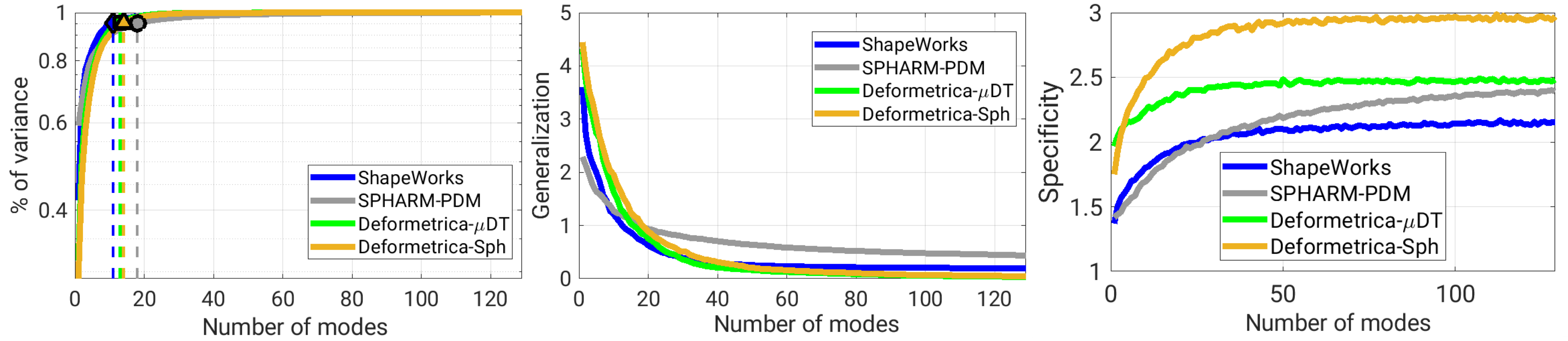}
	\caption{Compactness (higher is better), generalization (lower is better), and specificity (lower is better) analysis of LAA shape models}
	\label{fig:eval}
\end{figure}

\vspace{0.04in}
\noindent\textbf{Evaluation metrics}: Figure \ref{fig:eval} shows the quantitative metrics from each SSM tool in comparison. ShapeWorks produces the most compact shape model whereas SPHARM-PDM yielded the least compact one. Neither Deformetrica nor SPHARM-PDM attempt to learn the underlying shape distribution, their generalization was inferior with few prinicipal modes as compared to ShapeWorks, which reveals their tendency to overfit with sparse training samples in high-dimensional shape spaces. Furthermore, generated samples for ShapeWorks models preserved the shape characteristics of the given population, leading to a better specificity compared to shape models from the other tools. For Deformatrica models, the gap in the specificity performance indicated that selection of an atlas directly impacts the shape modeling process.

\begin{figure}[!htp]
	\centering%\vspace{-0.2in}
	\includegraphics[width=0.8\linewidth]{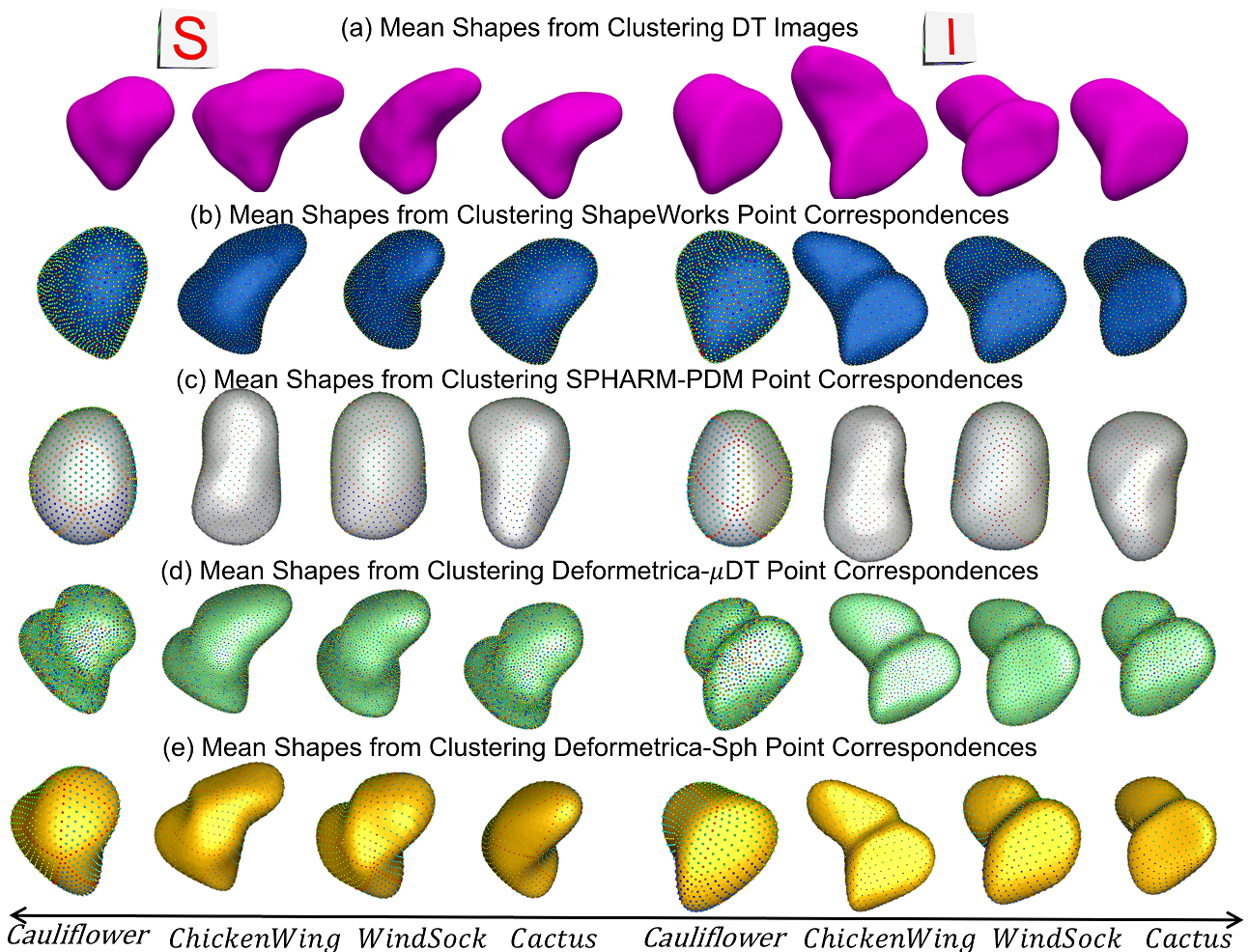}
	\caption{The mean shapes from k-means clustering (S: superior and I: inferior views) from (a) Distance Transform (DT) images, (b) ShapeWorks, (c) SPHARM-PDM, (d) Deformetrica-$\mu$DT, and (e) Deformetrica-Sph. Cluster centers from ShapeWorks and Deformetrica %shape
		 models are closely aligned with the centers from distance transforms.}%\vspace{-0.24in}
	\label{fig:clusters}
\end{figure}

%\vspace{-0.05in}
\noindent\textbf{Clustering analysis}: 
LAA has natural clustering in which the shapes are mainly categorized into four types: cauliflower, chicken wing, wind sock, and cactus \cite{wang2010}. 
K-means clustering was performed on the %point 
correspondences from each SSM tool to generate four %inherent 
clusters. For comparison, k-means was also performed on the preprocessed distance transforms.
The mean shapes of the four clusters from each tool were analyzed in comparison with the mean shapes of clustering from the distance transform images to explore the capability of the SSM tool to capture the intrinsic variability and the underlying clustering in the given population. Figure \ref{fig:clusters} demonstrates that cluster centers  from  ShapeWorks and Deformetrica shape models better matched those from distance transforms. % compared to SPHARM-PDM.

\begin{figure}[!htp]
	\centering%\vspace{-0.2in}
	\includegraphics[width=0.8\linewidth]{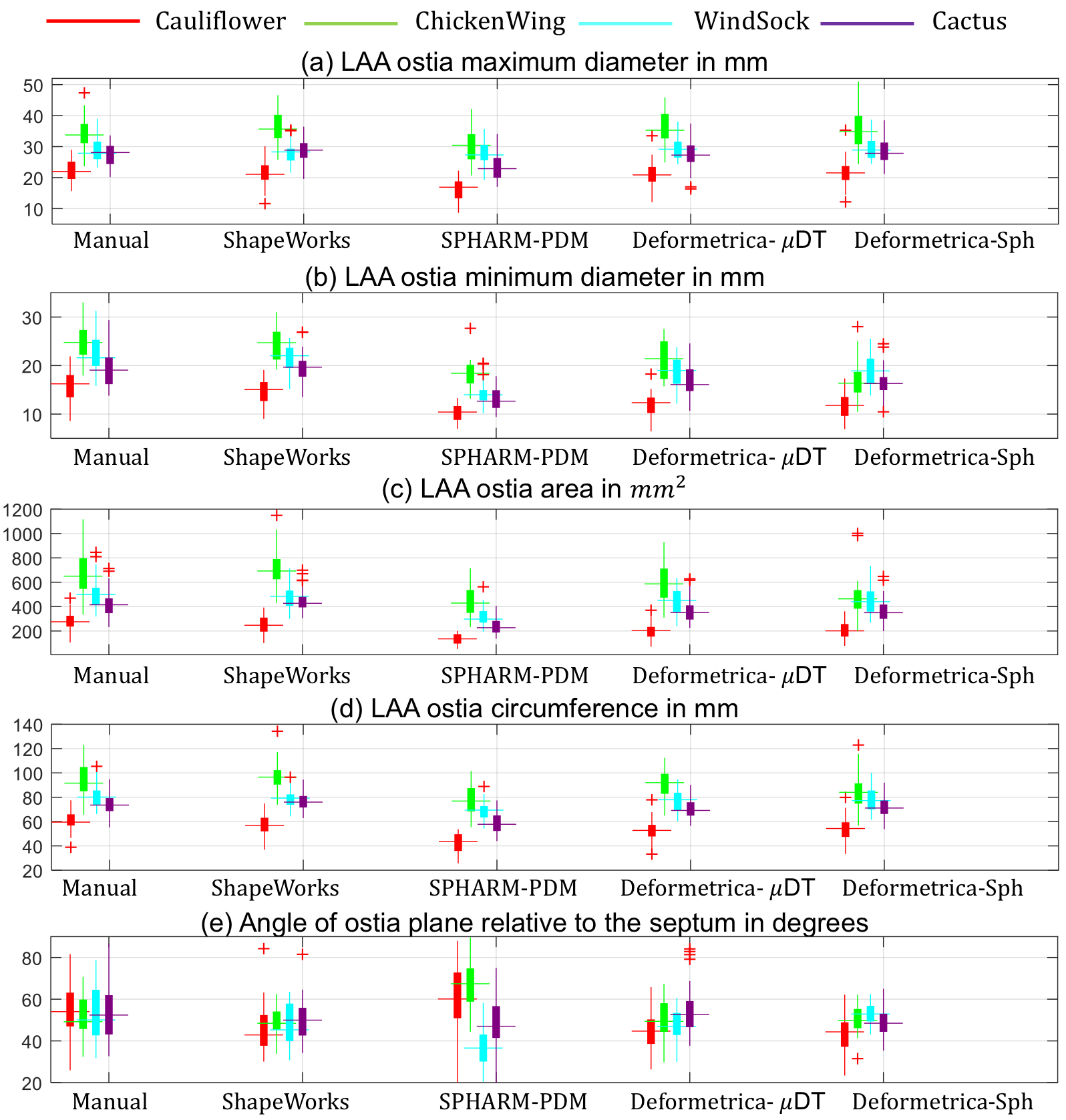}
	\caption{Box plots of LAA ostia measurements (a) maximum diameter in $mm$, (b) minimum diameter in $mm$, (c) area in $mm^2$, (d) circumference in $mm$, and (e) angle of ostia plane relative to the septum in degrees. 
	}%\vspace{-0.24in}
	\label{fig:measurements}
\end{figure}

%\FloatBarrier
\vspace{-0.15in}
\subsection{Shape models validation}
\vspace{-0.05in} 

For the 130 LAA shapes, the anatomical landmarks on the LAA ostia were obtained manually (and validated by a clinical expert) using Corview (Marrek inc., Salt Lake City, UT) to serve as a ground truth. 
%The landmarks are reviewed by a clinical expert. 
From the manual landmarks, the clinical measurements of LAA ostia such as minimum diameter, maximum diameter, area and circumference \cite{wang2010} were obtained by fitting an ellipse to the manual landmarks. To account for different LAA clusters, validation was performed by comparing measurements estimated based on shape models from SSM tools with the measurements estimated from manual landmarks  after clustering the preprocessed distance transforms into four clusters.
The measurements from the SSM tools were calculated semiautomatically by clustering the point correspondences from each tool into four categories. The mean shape from every cluster of each tool was used to manually mark the shape of the LAA ostia contour using Paraview \cite{ayachit2015paraview}. A clinical expert reviewed these contours. The point correspondences obtained on the ostium contour of the cluster mean shape were then warped back to the individual shape samples belonging to that cluster to generate the ostium contours of individual samples. % in the cluster of each tool using an automated approach. 
An ellipse was fit on the points on the ostium contour of each shape, and the measurements were computed. The manual and semiautomated measurements from each SSM tool were then compared to quantify the accuracy of the learned shape models in estimating the anatomical measurements. 
Figure \ref{fig:measurements} illustrates that the measurements obtained from ShapeWorks were the most consistent whereas those obtained from SPHARM-PDM were the least consistent.

\vspace{0.04in}
\noindent\textbf{Statistical testing}: The manual and semiautomated measurements from each SSM tool were compared using a paired t-test to identify if the differences were statistically significant. A paired t-test takes in measurements obtained by manual and semiautomated means and assumes a null hypothesis that both the recordings come from normal distributions with equal means and equal but unknown variances. The result of the test is rejected when the p-value$<$0.01 and is accepted when the p-value$>$0.01. Hence, for p-values$>$0.01, the measurement differences are not statistically significant. Tables 2-6 enumerate the p-values of a paired t-test for each measurement obtained from each tool's cluster in comparison to the manual measurements. Measurements obtained from ShapeWorks were more consistent as the p-values were always greater than 0.01. SPHARM-PDM was least consistent in most cases. 
The performance of Deformetrica for both the atlases was almost same for all the measurements. 

\begin{table}[H]
	\caption{%Paired t-test 
		p-values for LAA ostia maximum diameter measurement \vspace{-0.015in}}\label{tab1}
	\centering
	{\footnotesize
		\resizebox{0.7\linewidth}{!}{%
			\begin{tabular}{p{0.35\linewidth}p{0.17\linewidth}p{0.18\linewidth}p{0.16\linewidth}p{0.1\linewidth}} \toprule 
				\textbf{Comparison} & \textbf{Cauliflower} & \textbf{ChickenWing} & \textbf{WindSock} &\textbf{Cactus}\\
				\hline
				Manual vs ShapeWorks &  \textbf{p\textgreater{}0.01} & \textbf{p\textgreater{}0.01} & \textbf{p\textgreater{}0.01} & \textbf{p\textgreater{}0.01}\\
				Manual vs SPHARM-PDM &  {$p<0.01$} & \textbf{p\textgreater{}0.01} & \textbf{p\textgreater{}0.01} & {$p<0.01$}\\
				Manual vs Deformetrica-$\mu$DT &  \textbf{p\textgreater{}0.01} & \textbf{p\textgreater{}0.01} & \textbf{p\textgreater{}0.01} & \textbf{p\textgreater{}0.01}\\
				Manual vs Deformetrica-Sph &  \textbf{p\textgreater{}0.01} & \textbf{p\textgreater{}0.01}& \textbf{p\textgreater{}0.01} &\textbf{p\textgreater{}0.01}\\
				\hline
			\end{tabular}\vspace*{-\baselineskip}
		}
	}
\end{table}
%\FloatBarrier
%\FloatBarrier
\begin{table}[!htp]
	\caption{%Paired t-test 
		p-values for LAA ostia minimum diameter measurement \vspace{-0.015in}}\label{tab1}
	\centering
	{\footnotesize
		\resizebox{0.7\linewidth}{!}{%
			\begin{tabular}{p{0.35\linewidth}p{0.17\linewidth}p{0.18\linewidth}p{0.16\linewidth}p{0.1\linewidth}} \toprule 
				\textbf{Comparison} & \textbf{Cauliflower} & \textbf{ChickenWing} & \textbf{WindSock} &\textbf{Cactus}\\
				\hline
				Manual vs ShapeWorks &   \textbf{p\textgreater{}0.01} & \textbf{p\textgreater{}0.01} &  \textbf{p\textgreater{}0.01} &  \textbf{p\textgreater{}0.01}\\
				Manual vs SPHARM-PDM &  {$p<0.01$} &  \textbf{p\textgreater{}0.01} &  \textbf{p\textgreater{}0.01} & {$p<0.01$}\\
				Manual vs Deformetrica-$\mu$DT &  {$p<0.01$} & {$p<0.01$} & {$p<0.01$} & {$p<0.01$}\\
				Manual vs Deformetrica-Sph &  {$p<0.01$} & {$p<0.01$} & {$p<0.01$} & {$p<0.01$}\\
				\hline
			\end{tabular}\vspace*{-\baselineskip}
		}
	}
\end{table}
%\FloatBarrier
%\FloatBarrier
\begin{table}[!htp]
	\caption{%Paired t-test 
		p-values for LAA ostia area measurement \vspace{-0.015in}}\label{tab1}
	\centering
	{\footnotesize
		\resizebox{0.7\linewidth}{!}{%
			\begin{tabular}{p{0.35\linewidth}p{0.17\linewidth}p{0.18\linewidth}p{0.16\linewidth}p{0.1\linewidth}} \toprule 
				\textbf{Comparison} & \textbf{Cauliflower} & \textbf{ChickenWing} & \textbf{WindSock} &\textbf{Cactus}\\
				\hline
				Manual vs ShapeWorks &   \textbf{p\textgreater{}0.01} &  \textbf{p\textgreater{}0.01} &  \textbf{p\textgreater{}0.01} &  \textbf{p\textgreater{}0.01}\\
				Manual vs SPHARM-PDM &   {$p<0.01$} & {$p<0.01$} & {$p<0.01$} & {$p<0.01$}\\
				Manual vs Deformetrica-$\mu$DT &  {$p<0.01$} &  \textbf{p\textgreater{}0.01} & \textbf{p\textgreater{}0.01} & {$p<0.01$}\\
				Manual vs Deformetrica-Sph &  {$p<0.01$} & {$p<0.01$} &  \textbf{p\textgreater{}0.01} &  \textbf{p\textgreater{}0.01}\\
				\hline
			\end{tabular}\vspace*{-\baselineskip}
		}
	}
\end{table}
%\FloatBarrier
%\FloatBarrier
\begin{table}[!htp]
	\caption{%Paired t-test 
		p-values for LAA ostia circumference measurement\vspace{-0.015in}}\label{tab1}
	\centering
	{\footnotesize
		\resizebox{0.7\linewidth}{!}{%
			\begin{tabular}{p{0.35\linewidth}p{0.17\linewidth}p{0.18\linewidth}p{0.16\linewidth}p{0.1\linewidth}} \toprule 
				\textbf{Comparison} & \textbf{Cauliflower} & \textbf{ChickenWing} & \textbf{WindSock} &\textbf{Cactus}\\
				\hline
				Manual vs ShapeWorks &   \textbf{p\textgreater{}0.01} &  \textbf{p\textgreater{}0.01} &  \textbf{p\textgreater{}0.01} &  \textbf{p\textgreater{}0.01}\\
				Manual vs SPHARM-PDM &   {$p<0.01$} & {$p<0.01$} & {$p<0.01$} & {$p<0.01$}\\
				Manual vs Deformetrica-$\mu$DT &  {$p<0.01$} &  \textbf{p\textgreater{}0.01} & \textbf{p\textgreater{}0.01} & {$p<0.01$}\\
				Manual vs Deformetrica-Sph &  \textbf{p\textgreater{}0.01}&  \textbf{p\textgreater{}0.01} & \textbf{p\textgreater{}0.01} &  \textbf{p\textgreater{}0.01}\\
				\hline
			\end{tabular}\vspace*{-\baselineskip}
		}
	}
\end{table}
%\FloatBarrier
%\FloatBarrier
\begin{table}[!htp]
	\caption{%Paired t-test 
		p-values for angle of LAA ostia plane relative to septum measurement\vspace{-0.015in}}\label{tab1}
	\centering
	{\footnotesize
		\resizebox{0.7\linewidth}{!}{%
			\begin{tabular}{p{0.35\linewidth}p{0.17\linewidth}p{0.18\linewidth}p{0.16\linewidth}p{0.1\linewidth}} \toprule 
				\textbf{Comparison} & \textbf{Cauliflower} & \textbf{ChickenWing} & \textbf{WindSock} &\textbf{Cactus}\\
				\hline
				Manual vs ShapeWorks &   \textbf{p\textgreater{}0.01} &  \textbf{p\textgreater{}0.01} &  \textbf{p\textgreater{}0.01} &  \textbf{p\textgreater{}0.01}\\
				Manual vs SPHARM-PDM &   \textbf{p\textgreater{}0.01} & {$p<0.01$} & {$p<0.01$} &  \textbf{p\textgreater{}0.01}\\
				Manual vs Deformetrica-$\mu$DT &  {$p<0.01$} &  \textbf{p\textgreater{}0.01} &  \textbf{p\textgreater{}0.01} &  \textbf{p\textgreater{}0.01}\\
				Manual vs Deformetrica-Sph &  {$p<0.01$} &  \textbf{p\textgreater{}0.01} &  \textbf{p\textgreater{}0.01} &  \textbf{p\textgreater{}0.01}\\
				\hline
			\end{tabular}%\vspace*{-0.5\baselineskip}
		}
	}
\end{table}

%% file: conclusion.tex
\vspace{0.05in}
\section{Conclusion}
\vspace{-0.1in}

We presented an evaluation and a clinically driven validation framework for open-source statistical shape modeling (SSM) tools. SSM tools were evaluated quantitatively and qualitatively using measures such as compactness, generalization, specificity, modes of variation and intrinsic clustering discovery.  ShapeWorks and Deformetrica shape models were shown to capture clinically relevant population-level variability compared to SPHARM-PDM models. With lack of ground truth shape descriptors/correspondences, validating resulting shape models from different SSM tools becomes a challenge. To address this challenge, we have designed a semiautomated approach that is driven by learned shape models in the context of a clinical application to estimate clinically relevant anatomical measurements. 
Results emphasized the different levels of consistencies exhibited by different SSM tools. Yet, ShapeWorks -- by virtue of its optimized groupwise shape correspondences -- yields the most consistent anatomical measurements.
In the future, we will extend this study to other publicly available tools and clinical scenarios to benchmark SSM tools in different applications and to provide a blueprint for developing computational tools for shape models.